\documentclass[sigconf]{acmart}

\usepackage{booktabs}
\usepackage{multirow}

\usepackage{xcolor}
\newcommand{\AAc}[1]{\textcolor{green!60!black}{#1}}
\newcommand{\AAw}[1]{\textcolor{red}{#1}}

\usepackage{tikz}
\usepackage{pgfplots}
\pgfplotsset{compat=1.18} %

\AtBeginDocument{%
  }

\setcopyright{none}
\copyrightyear{2026}
\acmYear{2026}

\acmConference[KDD '26]{The 32nd ACM SIGKDD Conference on Knowledge Discovery and Data Mining}{August 9--13, 2026}{Jeju, Korea}

\begin{document}

\title[Refold]{Refold: Refining Protein Inverse Folding\\
with Efficient Structural Matching and Fusion}

\author{Yiran Zhu}
\affiliation{%
  \institution{AIMS Lab, HKUST (GZ)}
  \city{Guangzhou}
  \country{China}
}
\affiliation{%
  \institution{North China Electric Power University}
  \city{Baoding}
  \country{China}
}
\email{ciaran_study@yeah.net}

\author{Changxi Chi}
\affiliation{%
  \institution{Westlake University}
  \city{Hangzhou}
  \country{China}
}
\email{chichangxi@westlake.edu.cn}

\author{Hongxin Xiang}
\affiliation{%
  \institution{Hunan University}
  \city{Changsha}
  \country{China}
}
\email{xianghx@hnu.edu.cn}

\author{Wenjie Du}
\affiliation{%
  \institution{University of Science and Technology of China} %
  \city{Hefei}
  \country{China}
}
\email{duwenjie@mail.ustc.edu.cn}

\author{Xiaoqi Wang}
\affiliation{%
  \institution{Northwest Polytechnical University} %
  \city{Xi'an}
  \country{China}
}
\email{xqw@nwpu.edu.cn}

\author{Jun Xia}
\authornote{Corresponding author.}
\affiliation{%
  \institution{AIMS Lab, HKUST (GZ)}
  \city{Guangzhou}
  \country{China}
}
\affiliation{%
  \institution{HKUST}
  \city{Hong Kong SAR}
  \country{China}
}
\email{junxia@hkust-gz.edu.cn}

\renewcommand{\shortauthors}{Yiran Zhu et al.}

\begin{abstract}
Protein inverse folding aims to design an amino acid sequence that will fold into a given backbone structure, serving as a central task in protein design. Two main paradigms have been widely explored. Template-based methods exploit database-derived structural priors and can achieve high local precision when close structural neighbors are available, but their dependence on database coverage and match quality often degrades performance on out-of-distribution (OOD) targets. Deep learning approaches, in contrast, learn general structure-to-sequence regularities and usually generalize better to new backbones. However, they struggle to capture fine-grained local structure, which can cause uncertain residue predictions and missed local motifs in ambiguous regions. We introduce Refold, a novel framework that synergistically integrates the strengths of database-derived structural priors and deep learning prediction to enhance inverse folding. Refold obtains structural priors from matched neighbors and fuses them with model predictions to refine residue probabilities. In practice, low-quality neighbors can introduce noise, potentially degrading model performance. We address this issue with a Dynamic Utility Gate that controls prior injection and falls back to the base prediction when the priors are untrustworthy. Comprehensive evaluations on standard benchmarks demonstrate that Refold achieves state-of-the-art native sequence recovery of 0.63 on both CATH 4.2 and CATH 4.3. Also, analysis indicates that Refold delivers larger gains on high-uncertainty regions, reflecting the complementarity between structural priors and deep learning predictions. The code is available at \url{https://anonymous.4open.science/r/Refold-anon}.
\end{abstract}

\keywords{Protein inverse folding, Database-derived structural priors, Structural matching and fusion, Dynamic Utility Gate}

\maketitle

\section{Introduction}
\label{sec:intro1}

\begin{figure*}[t]
  \centering
  \includegraphics[width=0.85\linewidth]{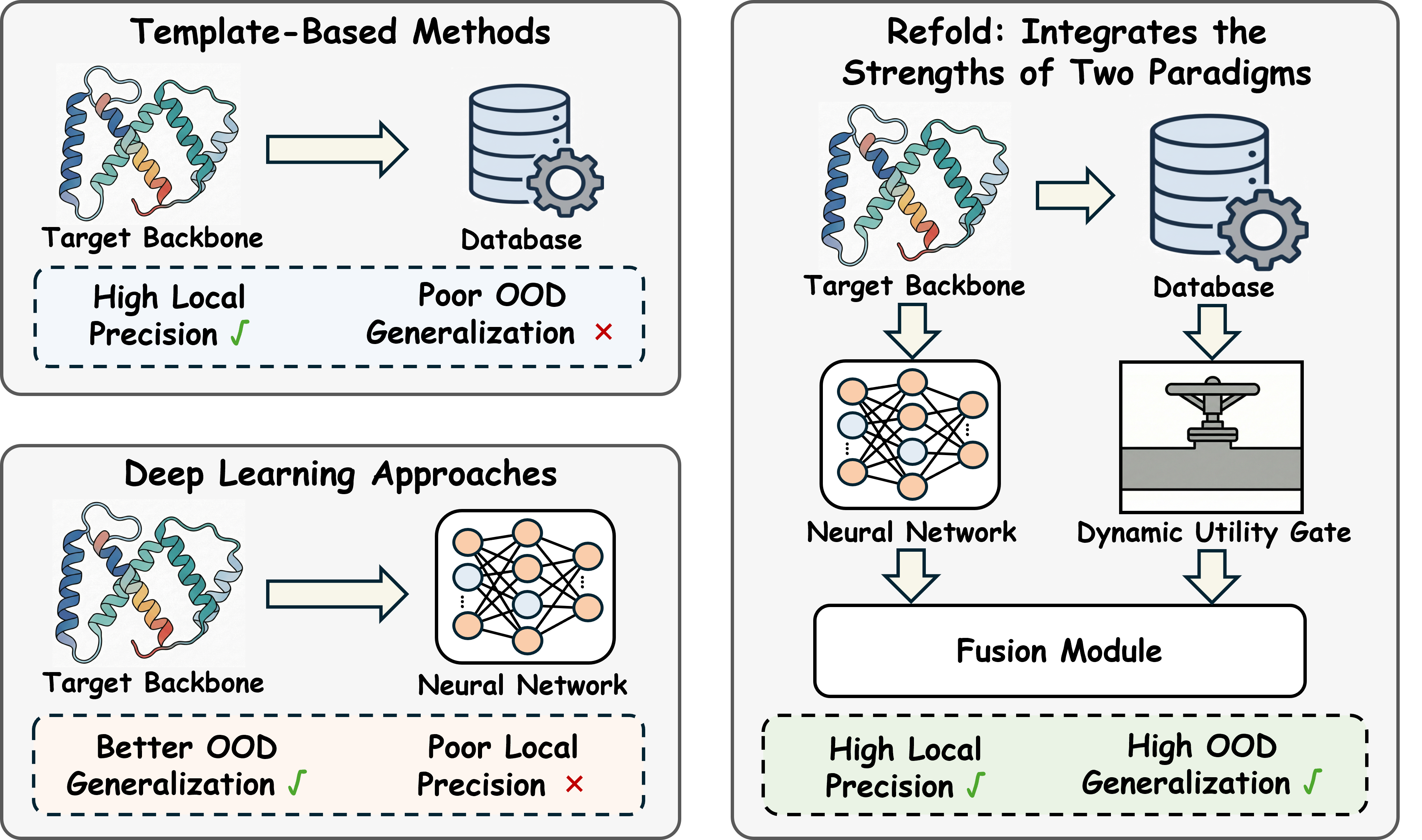}
  \caption{Template-based methods use database-derived structural priors to achieve high local precision but often show poor OOD generalization. Deep learning approaches generalize better to new backbones yet struggle with fine-grained local structure. Refold integrates these complementary strengths to improve both OOD generalization and local precision.}

  \label{fig:intro}
\end{figure*}

Designing amino acid sequences that fold into prescribed three-dimensional backbones is a foundational challenge in computational biology, commonly referred to as protein inverse folding. The task assigns an amino acid to each backbone position while satisfying physical and chemical constraints such as steric compatibility, hydrophobic interactions, and electrostatics. Accurate inverse folding is essential for understanding protein function and enabling protein design.

Traditional template-based methods rely on database-derived structural priors curated in large structural repositories such as the Protein Data Bank (PDB)~\cite{berman2000pdb, Sippl1995Knowledge}. Given a target backbone, they perform structural matching to retrieve similar folds or local fragments~\cite{Holm1993Dali} and transfer template-implied constraints (e.g., residue preferences, conserved motifs, or geometry-compatible configurations)~\cite{Sali1993Comparative}, often followed by packing/refinement under physics-inspired scoring. When matched structural neighbors exist, these priors can deliver high local precision. However, their dependence on database coverage and match quality can yield poor out-of-distribution (OOD) generalization when neighbors are absent or only remotely related, degrading designs for novel or weakly characterized folds. This limitation has motivated deep learning approaches that learn transferable structure-to-sequence regularities from large-scale data, albeit typically without explicitly leveraging matched-neighbor priors at inference time.

In recent years, deep learning has emerged as the dominant paradigm for protein inverse folding. Models such as ProteinMPNN \cite{dauparas2022robust}, PiFold \cite{gao2023pifold}, and KWDesign \cite{gao2024kwdesign} employ neural networks to model structure-to-sequence regularities from large-scale data. Through end-to-end training on diverse protein backbones, these methods achieve strong native sequence recovery and improved OOD generalization. However, lacking access to database-derived structural priors at inference time, these models are forced to depend entirely on learned patterns. This limitation becomes critical in structurally ambiguous regions, where weak local cues often lead to uncertain residue predictions and missed local motifs.

To unify the high local precision of template-based methods and the OOD generalization of deep learning approaches, we integrate these two paradigms into a framework, Refold. Given a query backbone, Refold performs efficient structural matching against a reference database to identify a set of matched neighbors, aligns them to the query, and derives structural priors. These priors are then fused with the model’s predictions to refine residue probabilities. Additionally, when the matched neighbors are of low quality, injecting these priors can overwrite correct predictions and negatively affect model performance. To address this issue, we introduce a Dynamic Utility Gate that adaptively controls the injection of structural priors and falls back to the original prediction when the priors are unreliable. As a result, Refold achieves state-of-the-art native sequence recovery on standard benchmarks, delivering substantial improvements particularly in structurally ambiguous regions where baseline models struggle. In addition, our framework is designed as a plug-and-play module that can be seamlessly integrated into diverse inverse folding base models. We highlight the core contributions of this work as follows.

\begin{itemize}
    \item As illustrated in Fig.~\ref{fig:intro}, we propose Refold, a plug-and-play framework that bridges the gap between template-based methods and deep learning approaches. By effectively integrating these complementary strengths, Refold significantly improves both local precision and OOD generalization.

    \item We propose an efficient structural matching-and-fusion algorithm that fuses the structural priors with model predictions to refine residue probabilities. We further introduce a Dynamic Utility Gate that adaptively controls the injection of structural priors to mitigate the adverse impact of low-quality matches.

    \item Extensive evaluations on standard benchmarks show that Refold delivers consistent gains and achieves state-of-the-art native sequence recovery.  As revealed in Sec.~\ref{sec:entropy_analysis}, the gains are most significant at high-uncertainty residues, empirically validating our solution to local structural ambiguities.

\end{itemize}

\begin{figure*}[t]
  \centering
  \includegraphics[width=0.9\linewidth]{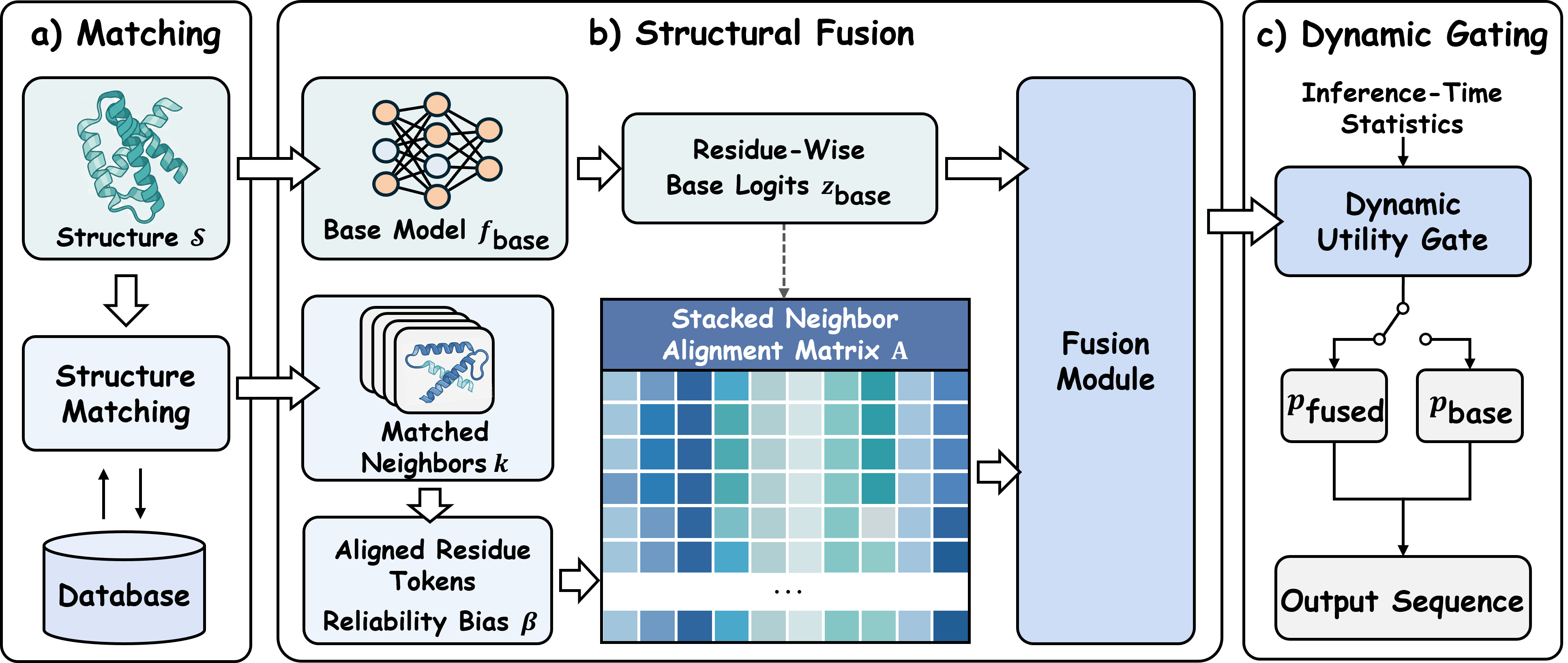}
  \caption{Overview of Refold. 
  (a) We perform structural matching to search matched neighbors $k$ for the target backbone $\mathcal{S}$. 
  (b) These neighbors are organized into a Stacked Neighbor Alignment matrix $\mathbf{A}$, where row 0 serves as a base anchor derived from the base logits $\operatorname*{argmax}(z_{\text{base}})$. 
  A Similarity-Weighted Fusion Module distills $\mathbf{A}$ into reference logits $z_{\text{ref}}$ and linearly fuses them with $z_{\text{base}}$ to form $p_{\text{fused}}$. 
  (c) To mitigate the adverse impact of low-quality neighbors, a Dynamic Utility Gate guided by the Inference-Time Statistics modulates the fusion, falling back to the base prediction when structural priors are unreliable.}
  \label{fig:method}
\end{figure*}

\section{Related Work} \label{sec:related_work}

\subsection{Template-Based Methods}
Early template-based pipelines leverage database-derived structural priors from the PDB~\cite{berman2000pdb} and perform structural matching to retrieve similar folds or local fragments~\cite{Holm1993Dali}, transferring template-implied constraints that guide sequence optimization on a fixed backbone. While they can achieve high local precision when close structural neighbors exist, their dependence on database coverage and match quality can limit generalization to out-of-distribution (OOD) backbones, where suitable matches are absent or only remotely related~\cite{Sali1993Comparative}.

\subsection{Deep Learning Approaches}
Recent data-driven approaches range from early autoregressive models like StructTrans~\cite{ingraham2019graphdesign} and GVP~\cite
{jing2021gvp} to state-of-the-art graph-based methods like ProteinMPNN~\cite{dauparas2022robust}, PiFold~\cite{gao2023pifold}, and KWDesign~\cite{gao2024kwdesign}.
While these models effectively capture global dependencies via message passing, they encode structural knowledge purely into model parameters.
Lacking inference-time access to matched neighbors, they often suffer from predictive uncertainty in regions with ambiguous local geometries.

\subsection{Utilization of Structural Priors}
Incorporating structural priors has proven effective in specialized domains, such as antibody design~\cite{wang2024retrievalab} and molecule generation~\cite{wang2025neural}.
However, extending this to general inverse folding is hindered by high structural diversity and noise.
Refold addresses this via a Dynamic Utility Gate that selectively integrates neighbor priors only when they provide reliable improvements, effectively bridging template-based precision with deep learning generalization.

\section{Method} 
\label{sec:method}

As illustrated in Fig.~\ref{fig:method}, Refold is a plug-and-play framework that augments base models via structural matching and fusion. It aligns database-derived priors with the query backbone and employs a Dynamic Utility Gate to selectively fuse these priors with base predictions, thereby preventing performance degradation from low-quality matches.

\subsection{Problem Formulation}
\label{sec:problem}

Protein inverse folding aims to design an amino acid sequence $\mathbf{y}=(y_1,\ldots,y_L)$ that is compatible with a prescribed 3D backbone structure $\mathcal{S}$ of length $L$.
Given $\mathcal{S}$, we seek to maximize the conditional likelihood $P(\mathbf{y}\mid \mathcal{S})$.

We build on a pre-trained inverse folding base model $f_{\text{base}}$ (e.g., ProteinMPNN), which operates on the geometric representation of $\mathcal{S}$ and outputs residue-wise base logits $z_{\text{base}}\in\mathbb{R}^{L\times 20}$.
We investigate two training strategies regarding the base model parameters: (1) Refold (Frozen), where we freeze $f_{\text{base}}$ throughout to strictly isolate the contribution of structural priors; and (2) Refold (Joint), where $f_{\text{base}}$ is jointly fine-tuned to maximize the synergistic integration for state-of-the-art performance. In both settings, we utilize the base output as the foundational prediction:
\begin{equation}
z_{\text{base}} = f_{\text{base}}(\mathcal{S}),
\end{equation}
where the logits correspond to the 20 canonical amino acids.
However, since these base predictions rely solely on learned parameters, structurally ambiguous regions often result in diffuse distributions with high entropy.
This characteristic reflects the predictive uncertainty of the base model, leading to suboptimal recovery of complex local motifs.
This limitation necessitates augmenting the base logits with structural priors derived from matched neighbors.

\begin{figure*}[t]
\centering
\includegraphics[width=0.9\linewidth]{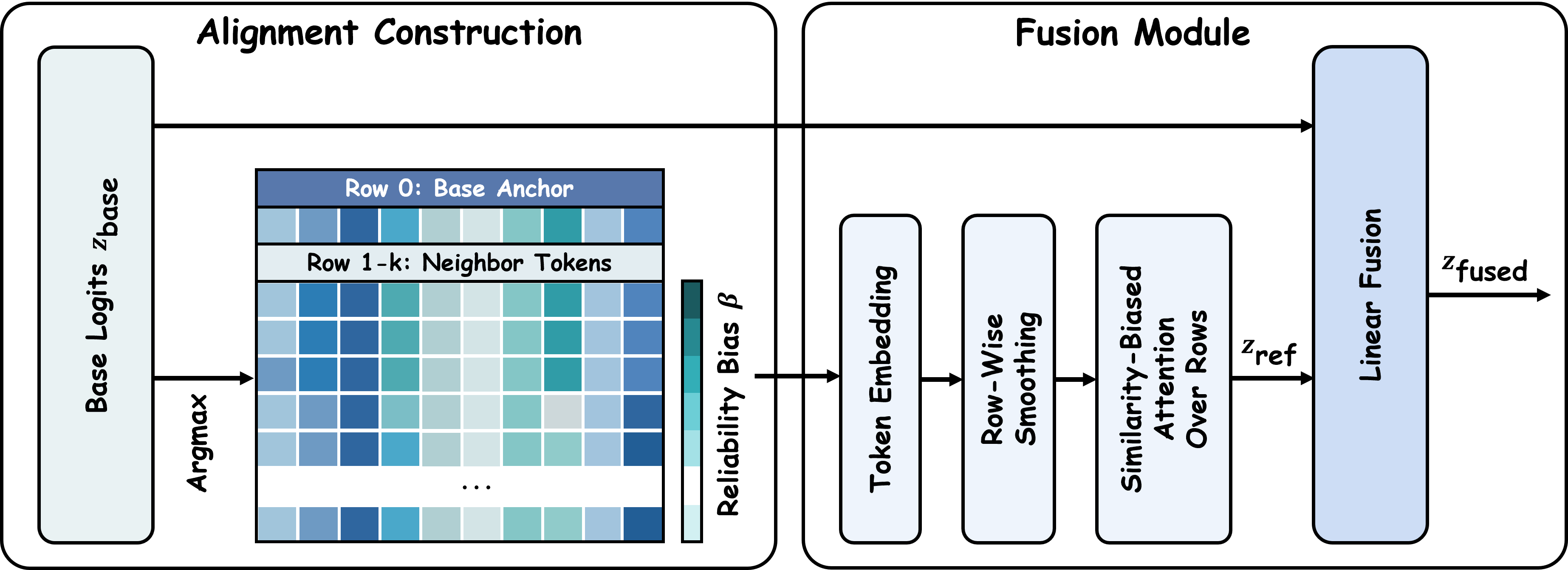} 
\caption{Schematic of the Similarity-Weighted Fusion Module. 
From the base logits $z_{\text{base}}$, we construct a Stacked Neighbor Alignment matrix $\mathbf{A}$: row~0 serves as the base anchor $\operatorname{argmax}\!\left(z_{\text{base}}\right)$, while rows $1,\dots,K$ contain aligned matched neighbor tokens. 
The module embeds $\mathbf{A}$, applies row-wise smoothing to mitigate local noise, and aggregates structural priors via attention weighted by the reliability bias $\beta$. 
This process produces reference logits $z_{\text{ref}}$, which are then residually fused with the base logits to yield $z_{\text{fused}} = z_{\text{base}} + \lambda \cdot z_{\text{ref}}$.}
\label{fig:fusion_detail}
\end{figure*}

\subsection{Constructing the Stacked Neighbor Alignment}
\label{sec:neighbor_alignment}
To augment the base model with structural priors, we search a reference database to find a set of matched neighbors for the query backbone $\mathcal{S}$.
We employ Foldseek \cite{van2023foldseek} to identify the top-$K$ structural candidates based on geometric similarity.
For each matched neighbor $k$, we generate a residue-level alignment mapping its amino acid sequence to the query indices $1,\dots,L$.
We further quantify the quality of each neighbor using a global similarity score $s_k$ (specifically, the TM-score).
To handle structural discrepancies, any query position that lacks a corresponding residue in the matched neighbor is filled with a special \texttt{GAP} token.

Matched neighbors often exhibit fragmented coverage and local misalignments.
We preserve their spatial details by organizing these priors into a \emph{Stacked Neighbor Alignment} matrix $\mathbf{A} \in \mathcal{V}^{(K+1) \times L}$.
Each column corresponds to a residue position on the query backbone.
Rows $r=1,\dots,K$ store the aligned residue tokens from the matched neighbors, while row $r=0$ serves as a base anchor.
Concretely, we instantiate this anchor via greedy decoding of the base logits:
\begin{equation}
y_{0,j} = \operatorname*{argmax}_{a \in \mathcal{V}} z_{\text{base},j}(a),
\end{equation}
where $\mathcal{V}$ denotes the set of 20 canonical amino acids.
This anchor provides a fully defined reference sequence, stabilizing downstream aggregation when structural priors are sparse or fragmented.
Consequently, each entry in $\mathbf{A}$ is a discrete token from $\mathcal{V}$ or \texttt{GAP}.

Recognizing that neighbor quality varies, we compute a reliability bias $\beta\in\mathbb{R}^{K+1}$ to guide the aggregation of structural priors.
Let $s=(s_1,\ldots,s_K)$ denote the global similarity scores (TM-scores) of the neighbors. We define:
\begin{equation}
\beta(r)=
\begin{cases}
\beta_0, & r=0,\\[4pt]
\mathrm{softmax}(s)_{r-1}, & r=1,\dots,K,
\end{cases}
\end{equation}
where $\beta_0$ is a learned scalar representing the baseline confidence.
In the subsequent Fusion Module, $\beta$ serves as a bias term in the attention mechanism.
This formulation encourages the model to prioritize high-similarity neighbors while retaining the capacity to reject irrelevant structural priors in favor of the base anchor.

\subsection{Similarity-Weighted Fusion Module}
\label{sec:fusion}

Given the Stacked Neighbor Alignment $\mathbf{A}$ and the reliability bias $\beta$ from Sec.~\ref{sec:neighbor_alignment}, we now describe a Similarity-Weighted Fusion Module designed to distill these structural priors into residue-wise reference logits $z_{\text{ref}}$.
These reference logits serve as a corrective signal that is subsequently combined with the base logits $z_{\text{base}}$, subject to modulation by the Dynamic Utility Gate described in the following subsection.
Fig.~\ref{fig:method} provides a schematic overview of this fusion process.

The module operates in three logical stages.
First, it spatially refines each neighbor row to encode local context and mitigate alignment noise.
Next, it aggregates structural priors across all neighbors at each query position, utilizing the reliability bias to weigh their contributions.
Finally, it projects the aggregated features into the target vocabulary space to produce $z_{\text{ref}}$ and integrates them with the base predictions via a residual connection.

\subsubsection{Row-wise Smoothing}
We first embed the discrete tokens in the alignment matrix $\mathbf{A}$ using a shared embedding layer $\mathbf{E}$, obtaining a dense tensor $X_0 \in \mathbb{R}^{(K+1) \times L \times d}$.
We then apply a depthwise-separable 1-D convolution along the sequence axis within each row independently, using a kernel size of 5 and a stride of 1.
Positions corresponding to \texttt{GAP} tokens are masked to ensure that missing residues do not propagate noise into the convolved features.
This row-wise smoothing aggregates short-range context within each neighbor sequence and helps mitigate small alignment offsets resulting from the structural matching process.

\subsubsection{Similarity-Biased Aggregation}
Next, we transpose the smoothed features to $X\in\mathbb{R}^{L\times (K+1)\times d}$ so that each target residue position $j$ can aggregate structural priors across the $K+1$ rows.
For each position, we compute multi-head attention over the neighbor dimension.
Queries, keys, and values are derived via standard linear projections of $X$.
\texttt{GAP} tokens in the key/value rows are masked to prevent the model from attending to missing data.
We augment the attention scores with the reliability bias $\beta$:
\begin{equation}
\ell_j(r\to r')=\frac{Q_j(r)\cdot K_j(r')}{\sqrt{d_h}}+\alpha \cdot \beta(r'),
\end{equation}
where $r'$ indexes the neighbor rows, $d_h$ is the per-head dimensionality, and $\alpha\ge 0$ is a learnable scaling factor.
The attention outputs are aggregated to produce $H\in\mathbb{R}^{L\times d}$, which is then projected through a feed-forward layer to the reference logits $z_{\text{ref}}\in\mathbb{R}^{L\times 20}$.

If a target position $j$ lacks valid neighbors (i.e., all tokens are \texttt{GAP}), we explicitly set $z_{\text{ref},j} = \mathbf{0}$, ensuring that the fusion branch is deactivated for that residue.

\subsubsection{Linear Fusion}
\label{sec:linear_fusion}
Finally, we integrate the structural priors with the base model's predictions via a linear combination of logits.
Let $z_{\text{base}}$ and $z_{\text{ref}}$ denote the residue-wise logits produced by the base model and the fusion module, respectively.

We define the fused logits $z_{\text{fused}}$ and the corresponding probability distribution $p_{\text{fused}}$ as:
\begin{equation}
z_{\text{fused}} = z_{\text{base}} + \lambda \cdot z_{\text{ref}}, \qquad
p_{\text{fused}} = \mathrm{softmax}(z_{\text{fused}}),
\end{equation}
where $\lambda$ is a learnable scalar controlling the strength of the reference signal.
This additive form acts as a residual refinement, allowing the robust base predictions to serve as a foundation while the reference path injects sparse, specific corrections.
When $z_{\text{ref}}$ provides little position-wise signal—for instance, when its magnitude is small or valid neighbors are absent—the fused distribution $p_{\text{fused}}$ naturally converges to the distribution implied by $z_{\text{base}}$.
In practice, this simple formulation is stable and avoids the additional assumptions required by more complex fusion schemes.

\subsection{Training and Robust Gated Inference}

\subsubsection{Training the Fusion Module}
We adopt a two-stage training strategy to ensure stability.
In the first stage, for Refold (Frozen), the base model $f_{\text{base}}$ remains frozen, and we optimize only the fusion module parameters $\theta$, including the shared embedding $\mathbf{E}$, the row-wise convolution, and the attention layers. For Refold (Joint), we unfreeze $f_{\text{base}}$ to enable end-to-end joint optimization of both the base model and fusion parameters.
The Dynamic Utility Gate is bypassed during this phase.
We minimize the cross-entropy loss between the fused distribution $p_{\text{fused}}$ and the ground-truth sequence $\mathbf{y}$:
\begin{equation}
\mathcal{L}(\theta) = -\sum_{i=1}^{L} \log p_{\text{fused}}(y_i\mid \mathcal{S}).
\end{equation}
Training on the full dataset encourages the fusion module to learn robust aggregation patterns even when structural priors are noisy or incomplete.

\subsubsection{Dynamic Utility Gate}
\label{sec:gate}

Indiscriminate fusion of structural priors can degrade performance when matched neighbors are globally similar but locally incompatible with the target.
To resolve this dilemma, we introduce a \emph{Dynamic Utility Gate} $G(\phi)$.
In the second training stage, we freeze both the base model and the fusion module, optimizing only the gate to predict whether the fused prediction outperforms the base model.

The gate input $\phi$ comprises inference-time statistics, including alignment coverage, the mean TM-score of matched neighbors, and divergence measures between the base distribution $p_{\text{base}}$ and the fused distribution $p_{\text{fused}}$.
We generate binary training labels $g \in \{0,1\}$ on a held-out validation set.
Let $\mathcal{L}_{\text{CE}}(p)$ denote the cross-entropy loss computed using distribution $p$.
We set $g=1$ if and only if $\mathcal{L}_{\text{CE}}(p_{\text{fused}}) < \mathcal{L}_{\text{CE}}(p_{\text{base}})$, and optimize $G(\phi)$ using binary cross-entropy.

At inference time, the gate functions as a global switch with a position-wise fallback.
If the predicted utility $G(\phi)$ falls below a threshold $\tau$ (tuned on validation data), the model reverts to the base distribution globally.
Otherwise, it adopts the fused distribution, defaulting to the base model only at specific positions where valid reference data is absent:
\begin{equation}
p_{\text{out},j}=
\begin{cases}
p_{\text{base},j}, & \text{if } G(\phi) < \tau \ \text{or } z_{\text{ref},j}=\mathbf{0},\\[4pt]
p_{\text{fused},j}, & \text{otherwise}.
\end{cases}
\end{equation}
This mechanism ensures that structural priors are utilized only when predicted to be beneficial, robustly defaulting to the base model when the gate rejects the fusion or when no aligned neighbors exist for a given residue. We summarize the inference cost of Refold (parameter and latency overhead) in Sec.~\ref{sec:sensitivity}.

\section{Experiments}
\label{sec:experiments}

\subsection{Experimental Settings}
\label{sec:settings}
In this work, leveraging the ProInvBench framework~\cite{gao2023proteininvbench}, we employ a non-redundant reference database alongside the CATH 4.2 and CATH 4.3 benchmarks~\cite{sillitoe2019cath42,sillitoe2021cath43}.
The datasets adhere to strict topology-based splitting protocols, ensuring robust data separation and preventing potential leakage between training and evaluation phases.
To further assess OOD generalization, we additionally include the TS50 and TS500 test sets~\cite{li2014direct}.

Refold is designed as a universal plug-and-play framework. We integrate the fusion module into ProteinMPNN~\cite{dauparas2022robust}, utilizing it as our primary base model.
We optimize the framework using the Adam optimizer with a linear warm-up schedule on an NVIDIA A800 GPU.
We train the model under two training variants (Frozen and Joint) to balance efficiency and performance.
For detailed hyperparameters and more training configurations, please refer to the Appendix \ref{app:datasets}.

\begin{table*}[t]
  \caption{Results comparison on the CATH dataset. ``Short'' refers to proteins with length $L < 100$, and ``Full'' corresponds to the complete test set. Sequence recovery is reported as a fraction (0--1). Values in parentheses indicate the absolute improvement (Recovery $\uparrow$) or reduction (Perplexity $\downarrow$) compared to the base model on the specific dataset version. Best results are bolded.}
  \label{tab:main_cath}

  \setlength{\tabcolsep}{13pt}
  \begin{tabular}{lcccccc}
    \toprule
    \multirow{2}{*}{Model} &
      \multicolumn{2}{c}{Perplexity $\downarrow$} &
      \multicolumn{2}{c}{Recovery $\uparrow$} &
      \multicolumn{2}{c}{CATH version} \\
    \cmidrule(lr){2-3} \cmidrule(lr){4-5} \cmidrule(lr){6-7}
      & Short & Full & Short & Full & 4.2 & 4.3 \\
    \midrule
    GraphTrans~\cite{ingraham2019graphdesign}  & 8.39 & 6.63 & 0.28 & 0.36 & $\checkmark$ & \\
    StructGNN~\cite{ingraham2019graphdesign}   & 8.29 & 6.40 & 0.29 & 0.36 & $\checkmark$ & \\
    ESM-IF$^\dagger$~\cite{hsu2022inversefolding}     & 8.18 & 6.44 & 0.31 & 0.38 &  & $\checkmark$ \\
    GCA~\cite{tan2022gca}                      & 7.09 & 6.05 & 0.33 & 0.38 & $\checkmark$ & \\
    GVP~\cite{jing2021gvp}                     & 7.23 & 5.36 & 0.31 & 0.39 & $\checkmark$ & \\
    GVP-large$^\dagger$~\cite{jing2021gvp}      & 7.68 & 6.17 & 0.32 & 0.39 &  & $\checkmark$ \\
    AlphaDesign~\cite{gao2022alphadesign}      & 7.32 & 6.30 & 0.34 & 0.41 & $\checkmark$ & \\
    PiFold~\cite{gao2023pifold}                                & 6.04 & 4.55 & 0.40 & 0.52 & $\checkmark$ & \\
    \midrule
    \multirow{2}{*}{ProteinMPNN~\cite{dauparas2022robust}}
      & 7.96 & 5.92 & 0.32 & 0.45 & $\checkmark$ & \\
      & 7.23 & 5.94 & 0.38 & 0.44 &  & $\checkmark$ \\
    \multirow{2}{*}{KWDesign~\cite{gao2024kwdesign}}
      & 5.48 & 3.46 & 0.44 & 0.61 & $\checkmark$ & \\
      & 5.47 & 3.49 & 0.51 & 0.60 &  & $\checkmark$ \\
    \midrule
    \multirow{2}{*}{\textbf{Refold (Frozen)}}
      & \textbf{5.38} {\scriptsize(-2.58)} & \textbf{3.84} {\scriptsize(-2.08)} & \textbf{0.49} {\scriptsize(+0.17)} & \textbf{0.61} {\scriptsize(+0.16)} & $\checkmark$ & \\
      & \textbf{4.61} {\scriptsize(-2.62)} & \textbf{3.77} {\scriptsize(-2.17)} & \textbf{0.53} {\scriptsize(+0.15)} & \textbf{0.61} {\scriptsize(+0.17)} &  & $\checkmark$ \\
    \multirow{2}{*}{\textbf{Refold (Joint)}}
      & \textbf{4.80} {\scriptsize(-3.16)} & \textbf{3.48} {\scriptsize(-2.44)} & \textbf{0.51} {\scriptsize(+0.19)} & \textbf{0.63} {\scriptsize(+0.18)} & $\checkmark$ & \\
      & \textbf{4.45} {\scriptsize(-2.78)} & \textbf{3.53} {\scriptsize(-2.41)} & \textbf{0.56} {\scriptsize(+0.18)} & \textbf{0.63} {\scriptsize(+0.19)} &  & $\checkmark$ \\
    \bottomrule
  \end{tabular}
\end{table*}

\subsection{Main Results}
\label{sec:main_results}

Table~\ref{tab:main_cath} compares performance on CATH 4.2 and 4.3. Refold achieves consistent improvements across all strategies.
First, Refold (Frozen) enhances the fixed ProteinMPNN, increasing recovery by +0.16/+0.17 on the full CATH 4.2/4.3 sets and reducing perplexity by 2.08/2.17.
The marked reduction in perplexity confirms that incorporating structural priors effectively mitigates the base model's predictive uncertainty.
Furthermore, Refold (Joint) establishes new state-of-the-art results via end-to-end optimization.
It attains 0.63 recovery on both benchmarks (+0.18/+0.19 gain), surpassing strong baselines like KWDesign and demonstrating the efficacy of fusing structural priors.

We further evaluate Refold on the TS50 and TS500 benchmarks, which provide structurally diverse targets under the standard protocol. As shown in Tab.~\ref{tab:generalization}, Refold yields consistent improvements on both TS50 and TS500, increasing recovery from 0.54 to 0.60 on TS50 and from 0.58 to 0.63 on TS500.
These gains under distribution shift suggest that structural priors from matched neighbors can effectively complement model predictions on structurally diverse targets.

\begin{table}[t]
  \caption{OOD Generalization performance on TS50 and TS500 test sets. We employ Refold (Frozen) trained on CATH~4.3. Best results are bolded.}
  \label{tab:generalization}
  \begin{tabular}{l l c c}
    \toprule
    Model & Test Set & Recovery $\uparrow$ & Perplexity $\downarrow$ \\
    \midrule
    \multirow{2}{*}{ProteinMPNN} 
      & TS50  & 0.54 & 3.93 \\
      & TS500 & 0.58 & 3.53 \\
    \midrule
    \multirow{2}{*}{\textbf{Refold}} 
      & TS50  & \textbf{0.60} {\scriptsize(+0.06)} & \textbf{3.78} {\scriptsize(-0.15)} \\
      & TS500 & \textbf{0.63} {\scriptsize(+0.05)} & \textbf{3.51} {\scriptsize(-0.02)} \\
    \bottomrule
  \end{tabular}
\end{table}

\subsection{Where Does Refold Help? Uncertainty Stratified Analysis}
\label{sec:entropy_analysis}

A central premise of Refold is that the base model tends to be unreliable in structurally ambiguous regions, where the backbone geometry supports multiple plausible residue identities.
In such scenarios, the base model typically yields a diffuse predictive distribution, characterized by high entropy.
Consequently, if Refold effectively leverages structural priors, we expect the performance gains to be concentrated in these high-uncertainty positions rather than being uniformly distributed.

To verify this, we compute the token-level Shannon entropy $H_j$ of the prediction at each residue position $j$.
Let $p_{\text{base}}(\cdot \mid j) = \mathrm{softmax}(z_{\text{base}, j})$ denote the probability distribution implied by the base logits. We define:
\begin{equation}
H_j \;=\; -\sum_{a \in \mathcal{V}} p_{\text{base}}(a \mid j)\,\log p_{\text{base}}(a \mid j),
\end{equation}
where $\mathcal{V}$ represents the set of 20 canonical amino acids.
We then stratify all validation tokens into three uncertainty regimes using fixed entropy thresholds: low uncertainty ($H_j < 1.5$), medium uncertainty ($1.5 \le H_j < 2.3$), and high uncertainty ($H_j \ge 2.3$).
These regimes contain 101{,}888, 82{,}763, and 87{,}067 tokens, respectively.

\begin{table}[t]
\caption{Uncertainty-stratified recovery analysis based on base model predictive entropy $H_j$. $\Delta$ denotes the absolute recovery gain. Note that gains are most pronounced in high-entropy regimes.}
\label{tab:entropy_gain}
\vskip 0.10in
\centering
\setlength{\tabcolsep}{6pt}
\renewcommand{\arraystretch}{1.1}
\begin{small}
\begin{sc}
\begin{tabular}{lccc}
\toprule
Uncertainty Regime & Base Rec. & Refold Rec. & $\Delta$ \\
\midrule
Low ($H<1.5$)        & 0.78 & 0.83 & +0.05 \\
Mid ($1.5\le H<2.3$) & 0.34 & 0.55 & +0.21 \\
High ($H\ge 2.3$)    & 0.18 & 0.45 & +0.27 \\
\midrule
Overall (All)        & 0.44 & 0.61 & +0.17 \\
\bottomrule
\end{tabular}
\end{sc}
\end{small}
\vskip -0.10in
\end{table}

\begin{figure*}[t]
  \centering
  \includegraphics[width=1\linewidth, trim=0.1in 0.8in 0.1in 0.5in, clip]{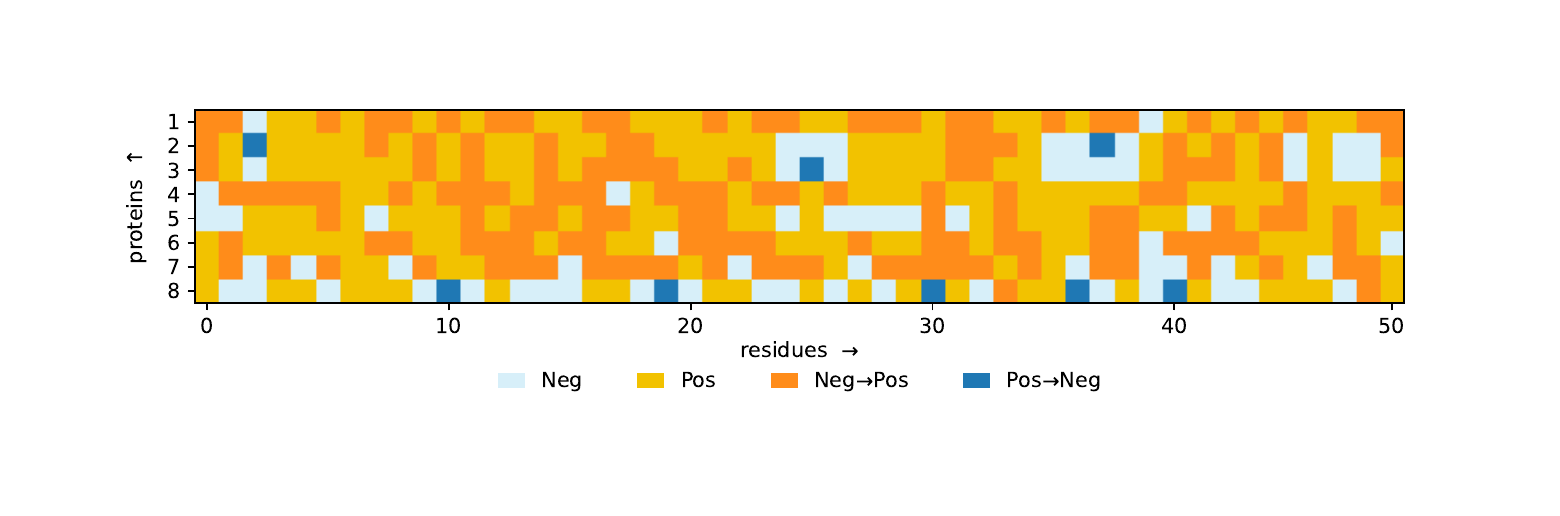}
  \caption{Site-wise recovery states (sampled proteins).
  We sample $N$ proteins and visualize the first 50 residues. Each cell denotes a residue-level transition from the Base model to Refold:
  Neg (wrong$\rightarrow$wrong), Pos (correct$\rightarrow$correct),
  Neg$\rightarrow$Pos (wrong$\rightarrow$correct, correction), and Pos$\rightarrow$Neg (correct$\rightarrow$wrong, error).
  Corrections (Neg$\rightarrow$Pos) tend to appear in localized segments, while error events (Pos$\rightarrow$Neg) are sparse, suggesting that Refold performs targeted, structure-aligned refinements rather than indiscriminate perturbations.}
  \label{fig:sitewise_states}
\end{figure*}

Table 3 reports the native sequence recovery across the stratified uncertainty regimes.
To strictly isolate the corrective effect of structural priors from base model adaptation, we conduct this analysis using Refold (Frozen) on CATH 4.3.
We observe that performance gains increase monotonically with the base model's uncertainty.
Specifically, Refold improves recovery by $+0.05$ on low-entropy tokens ($0.78 \rightarrow 0.83$), while the improvements are substantially amplified on mid-entropy ($+0.21$, $0.34 \rightarrow 0.55$) and high-entropy tokens ($+0.27$, $0.18 \rightarrow 0.45$).
This pattern corroborates our hypothesis that Refold primarily rectifies errors in structurally ambiguous regions where the base model is uncertain, utilizing structural priors as decisive corrections precisely when geometric reasoning alone is unreliable.
Overall, Refold improves token-level recovery from 0.44 to 0.61 ($+0.17$), consistent with the aggregate gains reported in the main results.

\subsection{Ablation Studies}
\label{sec:ablation}

We conduct ablation studies to validate the contribution of key components in Refold.
Consistent with Sec.~4.3, all ablations are performed using the Refold (Frozen) setting on CATH 4.3 to strictly isolate the impact of each module.

\begin{table}[t]
\caption{Ablation studies on ProteinMPNN and CATH 4.3.}
\label{tab:ablation}
\vskip 0.15in
\begin{center}
\begin{small}
\begin{sc}
\begin{tabular}{lcc}
\toprule
Configuration & Recovery & Perplexity \\
\midrule
\textbf{Full Model (Ours)} & \textbf{0.61} & \textbf{3.77} \\
\midrule
\multicolumn{3}{c}{Source Contribution} \\
\midrule
w/o Refold (Baseline)    & 0.44 & 5.94 \\
w/o base model (Priors Only) & 0.52 & 6.40 \\
\midrule
\multicolumn{3}{c}{Mechanisms} \\
\midrule
w/o TM-score Bias        & 0.58 & 4.21 \\
\midrule
\multicolumn{3}{c}{Utility Gate} \\
\midrule
w/o Gate                 & 0.58 & 4.20 \\
\midrule
\multicolumn{3}{c}{Aggregation Module} \\
\midrule
Row-only Attention       & 0.55 & 4.33 \\
Alternating Attention    & 0.50 & 4.92 \\
\bottomrule
\end{tabular}
\end{sc}
\end{small}
\end{center}
\vskip -0.1in
\end{table}

\begin{table*}[t]

\caption{
Token-level comparison of Base, Refold, and matched neighbor (Nbr) sequences for the 2kb9.A case.
Correct predictions are colored green, errors red, and Nbr positions highlighted in yellow indicate structure-guided corrections derived from the matched neighbor.
}
\label{tab:case_short}
\centering
\setlength{\tabcolsep}{2pt}
\renewcommand{\arraystretch}{1.2}

\begin{tabular}{l|*{30}{c}|c}
\toprule
Row &
0 & 1 & 2 & 3 & 4 & 5 & 6 & 7 & 8 & 9 &
10 & 11 & 12 & 13 & 14 & 15 & 16 & 17 & 18 & 19 &
20 & 21 & 22 & 23 & 24 & 25 & 26 & 27 & 28 & \dots \\
\midrule

Base &
\AAw{M} & \AAw{S} & \AAw{S} & \AAw{E} & \AAc{G} & \AAw{Y} & \AAw{S} & \AAc{G} & \AAw{E} & \AAw{S} &
\AAw{S} & \AAw{S} & \AAw{H} & \AAc{C} & \AAw{A} & \AAc{P} & \AAw{E} & \AAc{P} & \AAc{G} & \AAc{C} &
\AAw{T} & \AAw{R} & \AAc{G} & \AAw{T} & \AAw{S} & \AAw{S} & \AAw{R} & \AAc{P} & \AAw{R} & \dots \\

Refold &
\AAw{M} & \AAw{S} & \AAw{S} & \AAc{Y} & \AAc{G} & \AAc{W} & \AAw{S} & \AAc{G} & \AAc{L} & \AAw{S} &
\AAc{C} & \AAc{D} & \AAw{H} & \AAc{C} & \AAw{A} & \AAc{P} & \AAw{E} & \AAc{P} & \AAc{G} & \AAc{C} &
\AAw{T} & \AAw{R} & \AAc{G} & \AAw{T} & \AAc{C} & \AAw{S} & \AAw{R} & \AAc{P} & \AAc{R} & \dots \\

Nbr  &
\texttt{K} & \texttt{H} & \texttt{Q} & \colorbox{yellow!70}{\texttt{Y}} &
\texttt{G} & \colorbox{yellow!70}{\texttt{W}} & \texttt{Q} & \texttt{G} &
\colorbox{yellow!70}{\texttt{L}} & \texttt{Y} &
\colorbox{yellow!70}{\texttt{C}} & \colorbox{yellow!70}{\texttt{D}} &
\texttt{K} & \texttt{C} & \texttt{I} & \texttt{P} & \texttt{H} & \texttt{P} &
\texttt{G} & \texttt{C} &
\texttt{V} & \texttt{H} & \texttt{G} & \texttt{I} &
\colorbox{yellow!70}{\texttt{C}} & \texttt{N} & \texttt{E} & \texttt{P} &
\texttt{W} & \dots \\
\bottomrule
\end{tabular}

\end{table*}

\textit{Synergy of Geometric Reasoning and Structural Priors.}
We first investigate the individual contributions of the base model's geometric reasoning and the matched structural priors.
As shown in Tab.~\ref{tab:ablation}, relying solely on the base model (w/o Refold) yields a recovery of 0.44, whereas relying solely on structural priors (w/o Base Model) achieves 0.52.
Crucially, both single-source baselines fall significantly short of the full model (0.61).
This performance gap demonstrates that the two information sources are mutually complementary rather than redundant.
The structural matching and fusion process provides specific structural motifs that are difficult to predict from scratch, while the base model provides the necessary geometric reasoning to refine and contextualize these priors.
Refold achieves a true synergistic effect by effectively integrating these distinct information sources, ensuring that the fused output surpasses the capability of either individual source.

\textit{Efficacy of the Utility Gate.}
Comparing the full model to the ``w/o Gate'' variant, we verify the necessity of our Dynamic Utility Gate.
Removing the gate leads to a performance drop ($0.61 \to 0.58$), confirming that indiscriminate fusion of structural priors can hurt performance.
The gate effectively acts as a reliability filter, selectively suppressing low-confidence priors when they conflict with the robust base model.

\textit{Impact of Attention Mechanisms.}
Regarding aggregation, removing the TM-score bias leads to degradation (0.58), indicating that explicit similarity scores effectively guide the attention mechanism to prioritize high-quality neighbors.
Furthermore, our smoothed scheme outperforms naive Row-only or Alternating attention strategies, validating the design of Similarity-Weighted Fusion Module.

\subsection{Qualitative Analysis: Mining Local Geometric Motifs}
\label{sec:qualitative}

We further investigate the mechanism by which Refold leverages structural priors.
Since the base model relies solely on learned weights, it lacks access to specific structural priors at inference time.
When target regions exhibit subtle geometric ambiguity or weak local signals, the model depends entirely on its learned regularities, which often yields diffuse (high-entropy) residue distributions and consequently misses highly specific native motifs.

\textit{Sampled site-wise transition patterns.}
Before analyzing cases, we visualize token-level transitions on sampled proteins to inspect \emph{how} Refold alters predictions (Fig.~\ref{fig:sitewise_states}).
The transition map reveals that Refold's improvements often emerge as localized corrections (wrong$\rightarrow$correct), while correct$\rightarrow$wrong transitions remain rare.
This qualitative pattern supports our claim that Refold refines predictions in a targeted manner: it integrates structural priors where beneficial, without indiscriminately overwriting base predictions.

\textit{Case study: 2kb9.A.}
To demonstrate this granular mechanism, we analyze the test case 2kb9.A presented in Tab.~\ref{tab:case_short}.
For this target, the structural matching process identifies a matched neighbor with a TM-score of 0.757.

As visualized in the token-level comparison, the base model fails at several positions.
In these locally ambiguous regions, the base distribution becomes diffuse (high-entropy), indicating uncertainty and increasing the likelihood of missing specific native motifs.
In contrast, the matched neighbor, despite having a distinct global sequence, preserves precise local geometric motifs at these positions, carrying the correct residues Tyrosine (Y) and Cysteine (C).
Refold successfully attends to these structural priors (highlighted in yellow), correcting the prediction to the specific native residues.

This selective behavior confirms that Refold doesn't blindly copy the neighbor sequence; rather, it performs context-aware motif composition.
It leverages the base model's geometric reasoning to determine \emph{where} structural priors are reliable, and then integrates local priors to resolve high-entropy uncertainty into motif-specific predictions.

\begin{figure}[t]
  \centering
  \includegraphics[width=0.9\linewidth]{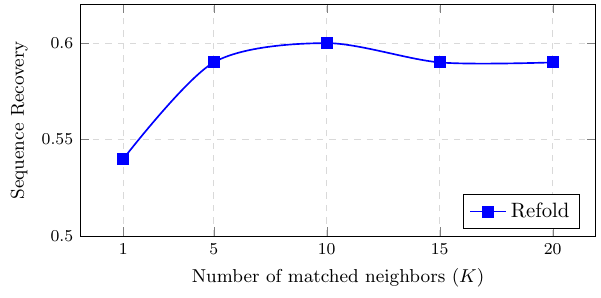}
  \caption{Sensitivity analysis of the number of matched neighbors $K$. The performance gain saturates around $K=10$, indicating that the model aggregates structural priors effectively without requiring a large number of neighbors.}
  \label{fig:sensitivity_k}
\end{figure}

\subsection{Sensitivity and Efficiency Analysis}
\label{sec:sensitivity}

We investigate the hyperparameter sensitivity and computational efficiency of Refold to assess its practical deployability.

\begin{table}[t]
\caption{Computational efficiency analysis on an NVIDIA A800 GPU. Latency is reported per protein with a batch size of 16. Overhead denotes the relative increase in latency compared to the base model ProteinMPNN}.
\label{tab:efficiency}
\vskip 0.15in
\begin{center}
\begin{small}
\begin{sc}
\begin{tabular}{lccc}
\toprule
Model & Params & Latency & Overhead \\
\midrule
ProteinMPNN & 1.7M & 13.41ms & -- \\
\textbf{Refold} & \textbf{3.5M} & \textbf{18.22ms} & \textbf{+35.9\%} \\
KWDesign & 4.1B & 656.24ms & +4793.7\% \\
\bottomrule
\end{tabular}
\end{sc}
\end{small}
\end{center}
\vskip -0.1in
\end{table}

\textit{Sensitivity to Neighbor Count ($K$).}
As illustrated in Fig.~\ref{fig:sensitivity_k}, we analyze the impact of the number of matched neighbors $K$. Sequence recovery improves rapidly as $K$ increases from 1 to 10 and plateaus thereafter. This suggests that the fusion module can extract sufficient structural context from a compact set of high-quality neighbors ($K \approx 10$), avoiding the computational overhead of processing large numbers of matches.

\textit{Inference Cost.}
We summarize parameter and latency overhead in Table~\ref{tab:efficiency}.
Refold introduces a minimal number of additional trainable parameters (approximately 1.8M) to the ProteinMPNN base model.
While this results in a moderate latency increase of $+35.9\%$ compared to the lightweight base model, the total inference time remains at the millisecond level (18.22ms).
In contrast, the previous state-of-the-art model KWDesign relies on a massive 4.1B parameter architecture, resulting in an inference latency of 656.24ms---over $30\times$ slower than our method.
These results demonstrate that Refold establishes a new state-of-the-art with orders of magnitude better efficiency than large-scale baselines, making it highly suitable for high-throughput applications.

\section{Conclusion}
In this paper, we presented Refold, which enhances protein inverse folding by fusing database-derived structural priors with deep learning predictions. Leveraging matched neighbors from a reference database, Refold improves both local precision and OOD generalization. Moreover, the Dynamic Utility Gate prevents performance degradation by avoiding over-reliance on noisy or locally incompatible neighbors, maintaining stable gains even when high-quality matched neighbors are scarce. As a result, Refold achieves state-of-the-art native sequence recovery on both CATH 4.2 and CATH 4.3. Looking ahead, future work could focus on more robust utility estimation to further reduce the risk of performance degradation, and on exploring alternative ways to integrate structural priors with model predictions.

\section{Limitations and Ethical Considerations}
\label{sec:limitations_ethics}

While Refold enhances inverse folding accuracy, the structural matching and fusion process adds a modest latency overhead ($+35.9\%$) compared to the lightweight base model, though it remains orders of magnitude faster than large-scale baselines. Additionally, our approach assumes a fixed backbone conformation, ignoring the dynamic nature of proteins in solution. Regarding ethical considerations, this work relies solely on public structural data (PDB) and involves no personal or sensitive information, though we acknowledge potential selection bias toward proteins that are easier to determine structurally. Finally, while intended to accelerate therapeutic design, we recognize the potential for dual-use misuse (e.g., harmful protein design); therefore, any generated sequences should be treated as computational hypotheses and undergo rigorous biosafety, toxicity, and institutional review before wet-lab synthesis.

\bibliographystyle{ACM-Reference-Format}
\bibliography{sample-base}

\end{document}